# Tensor Robust Principal Component Analysis: Exact Recovery of Corrupted Low-Rank Tensors via Convex Optimization

Canyi Lu[1], Jiashi Feng[1], Yudong Chen[2], Wei Liu[3], Zhouchen Lin[4,5,*] Shuicheng Yan[6,1]

[1] Department of Electrical and Computer Engineering, National University of Singapore
[2] School of Operations Research and Information Engineering, Cornell University    [3] Didi Research
[4] Key Laboratory of Machine Perception (MOE), School of EECS, Peking University
[5] Cooperative Medianet Innovation Center, Shanghai Jiaotong University    [6] 360 AI Institute
canyilu@gmail.com, elefjia@nus.edu.sg, yudong.chen@cornell.edu, wliu@ee.columbia.edu
zlin@pku.edu.cn, eleyans@nus.edu.sg

## Abstract

*This paper studies the Tensor Robust Principal Component (TRPCA) problem which extends the known Robust PCA [4] to the tensor case. Our model is based on a new tensor Singular Value Decomposition (t-SVD) [14] and its induced tensor tubal rank and tensor nuclear norm. Consider that we have a 3-way tensor $\mathcal{X} \in \mathbb{R}^{n_1 \times n_2 \times n_3}$ such that $\mathcal{X} = \mathcal{L}_0 + \mathcal{S}_0$, where $\mathcal{L}_0$ has low tubal rank and $\mathcal{S}_0$ is sparse. Is that possible to recover both components? In this work, we prove that under certain suitable assumptions, we can recover both the low-rank and the sparse components exactly by simply solving a convex program whose objective is a weighted combination of the tensor nuclear norm and the $\ell_1$-norm, i.e.,*

$$\min_{\mathcal{L},\mathcal{E}} \|\mathcal{L}\|_* + \lambda \|\mathcal{E}\|_1, \text{ s.t. } \mathcal{X} = \mathcal{L} + \mathcal{E},$$

*where $\lambda = 1/\sqrt{\max(n_1, n_2)n_3}$. Interestingly, TRPCA involves RPCA as a special case when $n_3 = 1$ and thus it is a simple and elegant tensor extension of RPCA. Also numerical experiments verify our theory and the application for the image denoising demonstrates the effectiveness of our method.*

## 1. Introduction

The problem of exploiting low-dimensional structure in high-dimensional data is taking on increasing importance in image, text and video processing, and web search, where the observed data lie in very high dimensional spaces. The classical Principal Component Analysis (PCA) [12] is the most widely used statistical tool for data analysis and dimensionality reduction. It is computationally efficient and powerful for the data which are mildly corrupted by small noises. However, a major issue of PCA is that it is brittle to grossly corrupted or outlying observations, which are ubiquitous in real world data. To date, a number of robust versions of PCA were proposed. But many of them suffer from the high computational cost.

---

[*]Corresponding author.

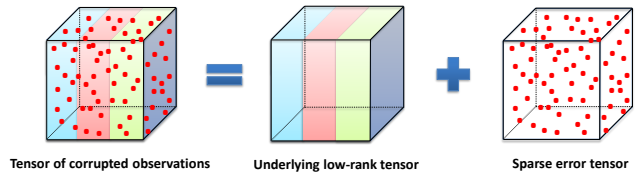

Figure 1: Illustration of the low-rank and sparse tensor decomposition from noisy observations.

The recent proposed Robust PCA [4] is the first polynomial-time algorithm with strong performance guarantees. Suppose we are given a data matrix $\boldsymbol{X} \in \mathbb{R}^{n_1 \times n_2}$, which can be decomposed as $\boldsymbol{X} = \boldsymbol{L}_0 + \boldsymbol{E}_0$, where $\boldsymbol{L}_0$ is low-rank and $\boldsymbol{E}_0$ is sparse. It is shown in [4] that if the singular vectors of $\boldsymbol{L}_0$ satisfy some incoherent conditions, $\boldsymbol{L}_0$ is low-rank and $\boldsymbol{S}_0$ is sufficiently sparse, then $\boldsymbol{L}_0$ and $\boldsymbol{S}_0$ can be recovered with high probability by solving the following convex problem

$$\min_{\boldsymbol{L},\boldsymbol{E}} \|\boldsymbol{L}\|_* + \lambda \|\boldsymbol{E}\|_1, \text{ s.t. } \boldsymbol{X} = \boldsymbol{L} + \boldsymbol{E}, \quad (1)$$

where $\|\boldsymbol{L}\|_*$ denotes the nuclear norm (sum of the singular values of $\boldsymbol{L}$), $\|\boldsymbol{E}\|_1$ denotes the $\ell_1$-norm (sum of the absolute values of all the entries in $\boldsymbol{E}$) and $\lambda = 1/\sqrt{\max(n_1, n_2)}$. RPCA and its extensions have been successfully applied for background modeling [4], video restoration [11], image alignment [22], et al.

One major shortcoming of RPCA is that it can only handle 2-way (matrix) data. However, the real world data are ubiquitously in multi-dimensional way, also referred to as

*tensor*. For example, a color image is a 3-way object with column, row and color modes; a greyscale video is indexed by two spatial variables and one temporal variable. To use RPCA, one has to first restructure/transform the multi-way data into a matrix. Such a preprocessing usually leads to the information loss and would cause performance degradation. To alleviate this issue, a common approach is to manipulate the tensor data by taking the advantage of its multi-dimensional structure.

In this work, we study the Tensor Robust Principal Component (TRPCA) which aims to exactly recover a low-rank tensor corrupted by sparse errors, see Figure 1 for an illustration. More specifically, suppose we are given a data tensor $\mathcal{X}$, and know that it can be decomposed as

$$\mathcal{X} = \mathcal{L}_0 + \mathcal{E}_0, \tag{2}$$

where $\mathcal{L}_0$ is low-rank and $\mathcal{E}_0$ is sparse; here, both components are of arbitrary magnitude. Note that we do not know the locations of the nonzero elements of $\mathcal{E}_0$, not even how many there are. Now we consider a similar problem as R-PCA. Can we recover the low-rank and sparse components exactly and efficiently from $\mathcal{X}$?

It is expected to extend the tools and analysis from the low-rank matrix recovery to the tensor case. However, this is not easy since the numerical algebra of tensors is fraught with hardness results [9]. The main issue for low-rank tensor estimation is the definition of tensor rank. Different from the matrix rank which enjoys several "good" properties, the tensor rank is not very well defined. Several different definitions of tensor rank have been proposed but each has its limitation. For example, the CP rank [15], defined as the smallest number of rank one tensor decomposition, is generally NP-hard to compute. Also its convex relaxation is intractable. Thus, the low CP rank tensor recovery is challenging. Another direction, which is more popular, is to use the tractable Tucker rank [15] and its convex relaxation. For a $k$-way tensor $\mathcal{X}$, the Tucker rank is a vector defined as $\text{rank}_{tc}(\mathcal{X}) := \left(\text{rank}\left(\boldsymbol{X}^{(1)}\right), \text{rank}\left(\boldsymbol{X}^{(2)}\right), \cdots, \text{rank}\left(\boldsymbol{X}^{(k)}\right)\right)$, where $\boldsymbol{X}^{(i)}$ is the mode-$i$ matricization of $\mathcal{X}$. The Tucker rank is based on the matrix rank and thus computable. Motivated from the fact that the nuclear norm is the convex envelop of the matrix rank within the unit ball of the spectral norm, the Sum of Nuclear Norms (SNN) [16], defined as $\sum_i \|\boldsymbol{X}^{(i)}\|_*$, is used as a convex surrogate of the Tucker rank. The effectiveness of this approach has been well studied in [16, 6, 26, 25]. However, SNN is not a tight convex relaxation of the Tucker rank [23]. The work [21] considers the low-rank tensor completion problem based on SNN,

$$\min_{\mathcal{X}} \sum_{i=1}^{k} \|\boldsymbol{X}^{(i)}\|_*, \text{ s.t. } P_{\boldsymbol{\Omega}}(\mathcal{X}) = P_{\boldsymbol{\Omega}}(\mathcal{X}_0), \tag{3}$$

where $P_{\boldsymbol{\Omega}}(\mathcal{X}_0)$ is an incomplete tensor with observed entries on the support $\boldsymbol{\Omega}$. It is shown in [21] that the above model can be substantially suboptimal: reliably recovering a $k$-way tensor of length $n$ and Tucker rank $(r, r, \cdots, r)$ from Gaussian measurements requires $O(rn^{k-1})$ observations. In contrast, a certain (intractable) nonconvex formulation needs only $O(rK + nrK)$ observations. The work [21] further proposes a better convexification based on a more balanced matricization of $\mathcal{X}$ and improves the bound to $O(r^{\lfloor \frac{k}{2} \rfloor} n^{\lfloor \frac{k}{2} \rfloor})$. It may be better than (3) for small $r$ and $k \geq 4$. But it is still far from optimal compared with the nonconvex model. Another work [10] proposes an SNN based tensor RPCA model

$$\min_{\mathcal{L}, \mathcal{E}} \sum_{i=1}^{k} \lambda_i \|\boldsymbol{L}^{(i)}\|_* + \|\mathcal{E}\|_1 + \frac{\tau}{2}\|\mathcal{L}\|_F^2 + \frac{\tau}{2}\|\mathcal{E}\|_F^2 \tag{4}$$
$$\text{s.t. } \mathcal{X} = \mathcal{L} + \mathcal{E}, \ \mathcal{X} \in \mathbb{R}^{n_1 \times n_2 \times \cdots \times n_k},$$

where $\|\mathcal{E}\|_1$ is the sum of the absolute values of all entries in $\mathcal{E}$, and gives the first exact recovery guarantee under certain tensor incoherence conditions.

More recently, the work [28] proposes the tensor tubal rank based on a new tensor decomposition scheme in [2, 14], which is referred as tensor SVD (t-SVD). The t-SVD is based on a new definition of tensor-tensor product which enjoys many similar properties as the matrix case. Based on the computable t-SVD, the tensor nuclear norm [24] is used to replace the tubal rank for low-rank tensor recovery (from incomplete/corrupted tensors) by solving the following convex program,

$$\min_{\mathcal{X}} \|\mathcal{X}\|_*, \text{ s.t. } P_{\boldsymbol{\Omega}}(\mathcal{X}) = P_{\boldsymbol{\Omega}}(\mathcal{X}_0), \tag{5}$$

where $\|\mathcal{X}\|_*$ denotes the tensor nuclear norm (see Section 2 for the definition).

In this work, we study the TRPCA problem which aims to recover the low tubal rank component $\mathcal{L}_0$ and sparse component $\mathcal{E}_0$ from $\mathcal{X} = \mathcal{L}_0 + \mathcal{E}_0 \in \mathbb{R}^{n_1 \times n_2 \times n_3}$ (this work focuses on the 3-way tensor) by convex optimization

$$\min_{\mathcal{L}, \mathcal{E}} \|\mathcal{L}\|_* + \lambda \|\mathcal{E}\|_1, \text{ s.t. } \mathcal{X} = \mathcal{L} + \mathcal{E}. \tag{6}$$

We prove that under certain incoherence conditions, the solution to the above problem perfectly recovers the low-rank and the sparse components, provided, of course that the tubal rank of $\mathcal{L}_0$ is not too large, and that $\mathcal{E}_0$ is reasonably sparse. A remarkable fact as that in TRPCA is that (6) has no tunning parameter either. Our analysis shows that $\lambda = 1/\sqrt{\max(n_1, n_2)n_3}$ guarantees the exact recovery no matter what $\mathcal{L}_0$ and $\mathcal{E}_0$ are. Actually, as will be seen later, if $n_3 = 1$ ($\mathcal{X}$ is a matrix in this case), our TRPCA in (6) reduces to RPCA in (1), and also our recovery guarantee in Theorem 3.1 reduces to Theorem 1.1 in [4]. Another advantage of (6) is that it can be solved by polynomial-time algorithms, e.g., the standard Alternating Direction Method of Multipliers (ADMM) [1].

The rest of this paper is organized as follows. Section 2 introduces some notations and preliminaries of tensors, where we define several algebraic structures of 3-way tensors. In Section 3, we will describe the main results of TRPCA and highlight some key differences from previous work. In Section 4, we conduct some experiments to verify our results in theory and apply TRPCA for image denoising. The last section gives concluding remarks and future directions.

## 2. Notations and Preliminaries

In this section, we introduce some notations and give the basic definitions used in the rest of the paper.

Throughout this paper, we denote tensors by boldface Euler script letters, e.g., $\mathcal{A}$. Matrices are denoted by boldface capital letters, e.g., $\boldsymbol{A}$; vectors are denoted by boldface lowercase letters, e.g., $\boldsymbol{a}$, and scalars are denoted by lowercase letters, e.g., $a$. We denote $\boldsymbol{I}_n$ as the $n \times n$ sized identity matrix. The filed of real number and complex number are denoted as $\mathbb{R}$ and $\mathbb{C}$, respectively. For a 3-way tensor $\mathcal{A} \in \mathbb{C}^{n_1 \times n_2 \times n_3}$, we denote its $(i,j,k)$-th entry as $\mathcal{A}_{ijk}$ or $a_{ijk}$ and use the Matlab notation $\mathcal{A}(i,:,:)$, $\mathcal{A}(:,i,:)$ and $\mathcal{A}(:,:,i)$ to denote respectively the $i$-th horizontal, lateral and frontal slice. More often, the frontal slice $\mathcal{A}(:,:,i)$ is denoted compactly as $\boldsymbol{A}^{(i)}$. The tube is denoted as $\mathcal{A}(i,j,:)$. The inner product of $\boldsymbol{A}$ and $\boldsymbol{B}$ in $\mathbb{C}^{n_1 \times n_2}$ is defined as $\langle \boldsymbol{A}, \boldsymbol{B} \rangle = \text{Tr}(\boldsymbol{A}^*\boldsymbol{B})$, where $\boldsymbol{A}^*$ denotes the conjugate transpose of $\boldsymbol{A}$ and $\text{Tr}(\cdot)$ denotes the matrix trace. The inner product of $\mathcal{A}$ and $\mathcal{B}$ in $\mathbb{C}^{n_1 \times n_2 \times n_3}$ is defined as $\langle \mathcal{A}, \mathcal{B} \rangle = \sum_{i=1}^{n_3} \langle \boldsymbol{A}^{(i)}, \boldsymbol{B}^{(i)} \rangle$.

Some norms of vector, matrix and tensor are used. We denote the $\ell_1$-norm as $\|\mathcal{A}\|_1 = \sum_{ijk} |a_{ijk}|$, the infinity norm as $\|\mathcal{A}\|_\infty = \max_{ijk} |a_{ijk}|$ and the Frobenius norm as $\|\mathcal{A}\|_F = \sqrt{\sum_{ijk} |a_{ijk}|^2}$. The above norms reduce to the vector or matrix norms if $\mathcal{A}$ is a vector or a matrix. For $\boldsymbol{v} \in \mathbb{C}^n$, the $\ell_2$-norm is $\|\boldsymbol{v}\|_2 = \sqrt{\sum_i |v_i|^2}$. The spectral norm of a matrix $\boldsymbol{A} \in \mathbb{C}^{n_1 \times n_2}$ is denoted as $\|\boldsymbol{A}\| = \max_i \sigma_i(\boldsymbol{A})$, where $\sigma_i(\boldsymbol{A})$'s are the singular values of $\boldsymbol{A}$. The matrix nuclear norm is $\|\boldsymbol{A}\|_* = \sum_i \sigma_i(\boldsymbol{A})$.

For $\mathcal{A} \in \mathbb{R}^{n_1 \times n_2 \times n_3}$, by using the Matlab command fft, we denote $\bar{\mathcal{A}}$ as the result of discrete Fourier transformation of $\mathcal{A}$ along the 3-rd dimension, i.e., $\bar{\mathcal{A}} = \texttt{fft}(\mathcal{A},[],3)$. In the same fashion, one can also compute $\mathcal{A}$ from $\bar{\mathcal{A}}$ via $\texttt{ifft}(\bar{\mathcal{A}},[],3)$ using the inverse FFT. In particular, we denote $\bar{\boldsymbol{A}}$ as a block diagonal matrix with each block on diagonal as the frontal slice $\bar{\boldsymbol{A}}^{(i)}$ of $\bar{\mathcal{A}}$, i.e.,

$$\bar{\boldsymbol{A}} = \texttt{bdiag}(\bar{\mathcal{A}}) = \begin{bmatrix} \bar{\boldsymbol{A}}^{(1)} & & & \\ & \bar{\boldsymbol{A}}^{(2)} & & \\ & & \ddots & \\ & & & \bar{\boldsymbol{A}}^{(n_3)} \end{bmatrix}.$$

The new tensor-tensor product [14] is defined based on an important concept, block circulant matrix, which can be regarded as a new matricization of a tensor. For $\mathcal{A} \in \mathbb{R}^{n_1 \times n_2 \times n_3}$, its block circulant matrix has size $n_1 n_3 \times n_2 n_3$, i.e.,

$$\texttt{bcirc}(\mathcal{A}) = \begin{bmatrix} \boldsymbol{A}^{(1)} & \boldsymbol{A}^{(n_3)} & \cdots & \boldsymbol{A}^{(2)} \\ \boldsymbol{A}^{(2)} & \boldsymbol{A}^{(1)} & \cdots & \boldsymbol{A}^{(3)} \\ \vdots & \vdots & \ddots & \vdots \\ \boldsymbol{A}^{(n_3)} & \boldsymbol{A}^{(n_3-1)} & \cdots & \boldsymbol{A}^{(1)} \end{bmatrix}.$$

We also define the following operator

$$\texttt{unfold}(\mathcal{A}) = \begin{bmatrix} \boldsymbol{A}^{(1)} \\ \boldsymbol{A}^{(2)} \\ \vdots \\ \boldsymbol{A}^{(n_3)} \end{bmatrix}, \ \texttt{fold}(\texttt{unfold}(\mathcal{A})) = \mathcal{A}.$$

Then t-product between two 3-way tensors can be defined as follows.

**Definition 2.1** *(t-product) [14] Let $\mathcal{A} \in \mathbb{R}^{n_1 \times n_2 \times n_3}$ and $\mathcal{B} \in \mathbb{R}^{n_2 \times l \times n_3}$. Then the t-product $\mathcal{A} * \mathcal{B}$ is defined to be a tensor of size $n_1 \times l \times n_3$,*

$$\mathcal{A} * \mathcal{B} = \texttt{fold}(\texttt{bcirc}(\mathcal{A}) \cdot \texttt{unfold}(\mathcal{B})). \quad (7)$$

Note that a 3-way tensor of size $n_1 \times n_2 \times n_3$ can be regarded as an $n_1 \times n_2$ matrix with each entry as a tube lies in the third dimension. Thus, the t-product is analogous to the matrix-matrix product except that the circular convolution replaces the product operation between the elements. Note that the t-product reduces to the standard matrix-matrix product when $n_3 = 1$. This is a key observation which makes our TRPCA involve RPCA as a special case.

**Definition 2.2** *(Conjugate transpose) [14] The conjugate transpose of a tensor $\mathcal{A}$ of size $n_1 \times n_2 \times n_3$ is the $n_2 \times n_1 \times n_3$ tensor $\mathcal{A}^*$ obtained by conjugate transposing each of the frontal slice and then reversing the order of transposed frontal slices 2 through $n_3$.*

**Definition 2.3** *(Identity tensor) [14] The identity tensor $\mathcal{I} \in \mathbb{R}^{n \times n \times n_3}$ is the tensor whose first frontal slice is the $n \times n$ identity matrix, and whose other frontal slices are all zeros.*

**Definition 2.4** *(Orthogonal tensor) [14] A tensor $\mathcal{Q} \in \mathbb{R}^{n \times n \times n_3}$ is orthogonal if it satisfies*

$$\mathcal{Q}^* * \mathcal{Q} = \mathcal{Q} * \mathcal{Q}^* = \mathcal{I}. \quad (8)$$

**Definition 2.5** *(F-diagonal Tensor) [14] A tensor is called f-diagonal if each of its frontal slices is a diagonal matrix.*

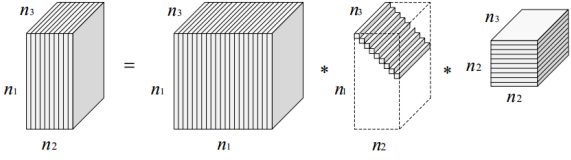

Figure 2: Illustration of the t-SVD of an $n_1 \times n_2 \times n_3$ tensor [13].

**Theorem 2.1** *(T-SVD) [14] Let $\mathcal{A} \in \mathbb{R}^{n_1 \times n_2 \times n_3}$. Then it can be factored as*

$$\mathcal{A} = \mathcal{U} * \mathcal{S} * \mathcal{V}^*, \quad (9)$$

*where $\mathcal{U} \in \mathbb{R}^{n_1 \times n_1 \times n_3}$, $\mathcal{V} \in \mathbb{R}^{n_2 \times n_2 \times n_3}$ are orthogonal, and $\mathcal{S} \in \mathbb{R}^{n_1 \times n_2 \times n_3}$ is a f-diagonal tensor.*

Figure 2 illustrates the t-SVD factorization. Note that t-SVD can be efficiently computed based on the matrix SVD in the Fourier domain. This is based on a key property that the block circulant matrix can be mapped to a block diagonal matrix in the Fourier domain, i.e.,

$$(\boldsymbol{F}_{n_3} \otimes \boldsymbol{I}_{n_1}) \cdot \text{bcirc}(\mathcal{A}) \cdot (\boldsymbol{F}_{n_3}^{-1} \otimes \boldsymbol{I}_{n_2}) = \bar{\boldsymbol{A}}, \quad (10)$$

where $\boldsymbol{F}_{n_3}$ denotes the $n_3 \times n_3$ Discrete Fourier Transform (DFT) matrix and $\otimes$ denotes the Kronecker product. Then one can perform the matrix SVD on each frontal slice of $\bar{\mathcal{A}}$, i.e., $\bar{\boldsymbol{A}}^{(i)} = \bar{\boldsymbol{U}}^{(i)} \bar{\boldsymbol{S}}^{(i)} \bar{\boldsymbol{V}}^{(i)*}$, where $\bar{\boldsymbol{U}}^{(i)}$, $\bar{\boldsymbol{S}}^{(i)}$ and $\bar{\boldsymbol{V}}^{(i)}$ are frontal slices of $\bar{\mathcal{U}}$, $\bar{\mathcal{S}}$ and $\bar{\mathcal{V}}$, respectively. Or equivalently, $\bar{\boldsymbol{A}} = \bar{\boldsymbol{U}} \bar{\boldsymbol{S}} \bar{\boldsymbol{V}}^*$. After performing ifft along the 3-rd dimension, we obtain $\mathcal{U} = \text{ifft}(\bar{\mathcal{U}}, [], 3)$, $\mathcal{S} = \text{ifft}(\bar{\mathcal{S}}, [], 3)$, and $\mathcal{V} = \text{ifft}(\bar{\mathcal{V}}, [], 3)$.

**Definition 2.6** *(Tensor multi rank and tubal rank) [28] The tensor **multi rank** of $\mathcal{A} \in \mathbb{R}^{n_1 \times n_2 \times n_3}$ is a vector $\boldsymbol{r} \in \mathbb{R}^{n_3}$ with its $i$-th entry as the rank of the $i$-th frontal slice of $\bar{\mathcal{A}}$, i.e., $r_i = \text{rank}(\bar{\boldsymbol{A}}^{(i)})$. The tensor **tubal rank**, denoted as $\text{rank}_t(\mathcal{A})$, is defined as the number of nonzero singular tubes of $\mathcal{S}$, where $\mathcal{S}$ is from the t-SVD of $\mathcal{A} = \mathcal{U} * \mathcal{S} * \mathcal{V}^*$. That is*

$$\text{rank}_t(\mathcal{A}) = \#\{i : \mathcal{S}(i,i,:) \neq \mathbf{0}\} = \max_i r_i. \quad (11)$$

The tensor tubal rank has some interesting properties as the matrix rank, e.g., for $\mathcal{A} \in \mathbb{R}^{n_1 \times n_2 \times n_3}$, $\text{rank}_t(\mathcal{A}) \leq \min(n_1, n_2)$, and $\text{rank}_t(\mathcal{A} * \mathcal{B}) \leq \min(\text{rank}_t(\mathcal{A}), \text{rank}_t(\mathcal{B}))$.

**Definition 2.7** *(Tensor nuclear norm) The tensor nuclear norm of a tensor $\mathcal{A} \in \mathbb{R}^{n_1 \times n_2 \times n_3}$, denoted as $\|\mathcal{A}\|_*$, is defined as the average of the nuclear norm of all the frontal slices of $\bar{\mathcal{A}}$, i.e., $\|\mathcal{A}\|_* := \frac{1}{n_3} \sum_{i=1}^{n_3} \|\bar{\boldsymbol{A}}^{(i)}\|_*$.*

With the factor $1/n_3$, our tensor nuclear norm definition is different from previous work [24, 28]. Note that this is important for our model and analysis in theory. The above tensor nuclear norm is defined in the Fourier domain. It is closely related to the nuclear norm of the block circulant matrix in the original domain. Indeed,

$$\|\mathcal{A}\|_* = \frac{1}{n_3} \sum_{i=1}^{n_3} \|\bar{\boldsymbol{A}}^{(i)}\|_* = \frac{1}{n_3} \|\bar{\boldsymbol{A}}\|_*$$
$$= \frac{1}{n_3} \|(\boldsymbol{F}_{n_3} \otimes \boldsymbol{I}_{n_1}) \cdot \text{bcirc}(\mathcal{A}) \cdot (\boldsymbol{F}_{n_3}^{-1} \otimes \boldsymbol{I}_{n_2})\|_* \quad (12)$$
$$= \frac{1}{n_3} \|\text{bcirc}(\mathcal{A})\|_*.$$

The above relationship gives an equivalent definition of the tensor nuclear norm in the original domain. So the tensor nuclear norm is the nuclear norm (with a factor $1/n_3$) of a new matricization (block circulant matrix) of a tensor. Compared with previous matricizations along certain dimension, the block circulant matricization may preserve more spacial relationship among entries.

**Definition 2.8** *(Tensor spectral norm) The tensor spectral norm of $\mathcal{A} \in \mathbb{R}^{n_1 \times n_2 \times n_3}$, denoted as $\|\mathcal{A}\|$, is defined as $\|\mathcal{A}\| := \|\bar{\boldsymbol{A}}\|$.*

If we further define the tensor average rank as $\text{rank}_a(\mathcal{A}) = \frac{1}{n_3} \sum_{i=1}^{n_3} \text{rank}(\bar{\boldsymbol{A}}^{(i)})$, then it can be proved that the tensor nuclear norm is the convex envelop of the tensor average rank within the unit ball of the tensor spectral norm.

**Definition 2.9** *(Standard tensor basis) [27] The **column basis**, denoted as $\mathring{\mathfrak{e}}_i$, is a tensor of size $n \times 1 \times n_3$ with its $(i,1,1)$-th entry equaling to 1 and the rest equaling to 0. Naturally its transpose $\mathring{\mathfrak{e}}_i^*$ is called **row basis**. The **tube basis**, denoted as $\dot{\mathfrak{e}}_k$, is a tensor of size $1 \times 1 \times n_3$ with its $(1,1,k)$-entry equaling to 1 and the rest equaling to 0.*

For simplicity, denote $\mathfrak{e}_{ijk} = \mathring{\mathfrak{e}}_i * \dot{\mathfrak{e}}_k * \mathring{\mathfrak{e}}_j^*$. Then for any $\mathcal{A} \in \mathbb{R}^{n_1 \times n_2 \times n_3}$, we have $\mathcal{A} = \sum_{ijk} \langle \mathfrak{e}_{ijk}, \mathcal{A} \rangle \mathfrak{e}_{ijk} = \sum_{ijk} a_{ijk} \mathfrak{e}_{ijk}$.

## 3. Tensor RPCA and Our Results

As in low-rank matrix recovery problems, some incoherence conditions need to be met if recovery is to be possible for tensor-based problems. Hence, in this section, we first introduce some incoherence conditions of the tensor $\mathcal{L}_0$ extended from [27, 4]. Then we present the recovery guarantee of TRPCA (6).

### 3.1. Tensor Incoherence Conditions

As observed in RPCA [4], the exact recovery is impossible in some cases. TRPCA suffers from a similar issue.

For example, suppose $\mathcal{X} = \mathring{\mathfrak{e}}_1 * \dot{\mathfrak{e}}_1 * \mathring{\mathfrak{e}}_1^*$ ($x_{ijk} = 1$ when $i = j = k = 1$ and zeros everywhere else). Then $\mathcal{X}$ is both low-rank and sparse. We are not able to identify the low-rank component and the sparse component in this case. To avoid such pathological situations, we need to assume that the low-rank component $\mathcal{L}_0$ is not sparse. To this end, we assume $\mathcal{L}_0$ to satisfy some incoherence conditions.

**Definition 3.1** *(Tensor Incoherence Conditions) For $\mathcal{L}_0 \in \mathbb{R}^{n_1 \times n_2 \times n_3}$, assume that $\text{rank}_t(\mathcal{L}_0) = r$ and it has the skinny t-SVD $\mathcal{L}_0 = \mathcal{U} * \mathcal{S} * \mathcal{V}^*$, where $\mathcal{U} \in \mathbb{R}^{n_1 \times r \times n_3}$ and $\mathcal{V} \in \mathbb{R}^{n_2 \times r \times n_3}$ satisfy $\mathcal{U}^* * \mathcal{U} = \mathcal{I}$ and $\mathcal{V}^* * \mathcal{V} = \mathcal{I}$, and $\mathcal{S} \in \mathbb{R}^{r \times r \times n_3}$ is a f-diagonal tensor. Then $\mathcal{L}_0$ is said to satisfy the tensor incoherence conditions with parameter $\mu$ if*

$$\max_{i=1,\cdots,n_1} \|\mathcal{U}^* * \mathring{\mathfrak{e}}_i\|_F \leq \sqrt{\frac{\mu r}{n_1 n_3}}, \tag{13}$$

$$\max_{j=1,\cdots,n_2} \|\mathcal{V}^* * \mathring{\mathfrak{e}}_j\|_F \leq \sqrt{\frac{\mu r}{n_2 n_3}}, \tag{14}$$

*and*

$$\|\mathcal{U} * \mathcal{V}^*\|_\infty \leq \sqrt{\frac{\mu r}{n_1 n_2 n_3^2}}. \tag{15}$$

As discussed in [4, 3], the incoherence condition guarantees that for small values of $\mu$, the singular vectors are reasonably spread out, or not sparse. As observed in [5], the joint incoherence condition is not necessary for low-rank matrix completion. However, for RPCA, it is unavoidable for polynomial-time algorithms. In our proofs, the joint incoherence (15) condition is necessary.

Another identifiability issue arises if the sparse tensor has low tubal rank. This can be avoided by assuming that the support of $\mathcal{S}_0$ is uniformly distributed.

### 3.2. Main Results

Now we show that, the convex program (6) is able to perfectly recover the low-rank and sparse components. We define $n_{(1)} = \max(n_1, n_2)$ and $n_{(2)} = \min(n_1, n_2)$.

**Theorem 3.1** *Suppose $\mathcal{L}_0 \in \mathbb{R}^{n \times n \times n_3}$ obeys (13)-(15). Fix any $n \times n \times n_3$ tensor $\mathcal{M}$ of signs. Suppose that the support set $\Omega$ of $\mathcal{S}_0$ is uniformly distributed among all sets of cardinality $m$, and that $\text{sgn}([\mathcal{S}_0]_{ijk}) = [\mathcal{M}]_{ijk}$ for all $(i,j,k) \in \Omega$. Then, there exist universal constants $c_1, c_2 > 0$ such that with probability at least $1 - c_1(nn_3)^{-c_2}$ (over the choice of support of $\mathcal{S}_0$), $\{\mathcal{L}_0, \mathcal{S}_0\}$ is the unique minimizer to (6) with $\lambda = 1/\sqrt{nn_3}$, provided that*

$$\text{rank}_t(\mathcal{L}_0) \leq \frac{\rho_r n n_3}{\mu(\log(nn_3))^2} \text{ and } m \leq \rho_s n^2 n_3, \tag{16}$$

*where $\rho_r$ and $\rho_s$ are positive constants. If $\mathcal{L}_0 \in \mathbb{R}^{n_1 \times n_2 \times n_3}$ has rectangular frontal slices, TRPCA with $\lambda = 1/\sqrt{n_{(1)} n_3}$ succeeds with probability at least $1 - c_1(n_{(1)} n_3)^{-c_2}$, provided that $\text{rank}_t(\mathcal{L}_0) \leq \frac{\rho_r n_{(2)} n_3}{\mu(\log(n_{(1)} n_3))^2}$ and $m \leq \rho_s n_1 n_2 n_3$.*

The above result shows that for incoherent $\mathcal{L}_0$, the perfect recovery is guaranteed with high probability for $\text{rank}_t(\mathcal{L}_0)$ on the order of $nn_3/(\mu(\log nn_3)^2)$ and a number of nonzero entries in $\mathcal{S}_0$ on the order of $n^2 n_3$. For $\mathcal{S}_0$, we make only an assumption on the random location distribution, but no assumption about the magnitudes or signs of the nonzero entries. Also TRPCA is parameter free. Moreover, since the t-product of 3-way tensors reduces to the standard matrix-matrix product when the third dimension is 1, the tensor nuclear norm reduces to the matrix nuclear norm. Thus, RPCA is a special case of TRPCA and the guarantee of R-PCA in Theorem 1.1 in [4] is a special case of our Theorem 3.1. Both our model and theoretical guarantee are consistent with RPCA. Compared with the SNN model [10], our tensor extension of RPCA is much more simple and elegant.

It is worth mentioning that this work focuses on the analysis for 3-way tensors. But it may not be difficult to generalize our model in (6) and results in Theorem 3.1 to the case of order-$p$ ($p \geq 3$) tensors, by using the t-SVD for order-$p$ tensors in [19].

### 3.3. Optimization by ADMM

ADMM is the most widely used solver for RPCA and its related problems. The work [28] also uses ADMM to solve a similar problem as (6). In this work, we also use ADMM to solve (6) and give the details here since the setting of some parameters are different but critical in the experiments. See Algorithm 1 for the optimization details. Note that both the updates of $\mathcal{L}_{k+1}$ and $\mathcal{S}_{k+1}$ have closed form solutions [28]. It is easy to see that the main per-iteration cost lies in the update of $\mathcal{L}_{k+1}$, which requires computing FFT and $n_3$ SVDs of $n_1 \times n_2$ matrices. Thus the per-iteration complexity is $O\left(n_1 n_2 n_3 \log(n_3) + n_{(1)} n_{(2)}^2 n_3\right)$.

## 4. Experiments

In this section, we conduct numerical experiments to corroborate our main results[1]. We first investigate the ability of TRPCA for recovering tensors of various tubal rank from noises of various sparsity. We then apply TRPCA for image denoising. Note that the choice of $\lambda$ in (6) is critical for the recovery performance. To verify the correctness of our main results, we set $\lambda = 1/\sqrt{n_{(1)} n_3}$ in all the experiments. But note that it is possible to further improve the performance by tuning $\lambda$ more carefully. The suggested value in theory provides a good guide in practice.

---

[1]Codes: https://github.com/canyilu/LibADMM

**Algorithm 1** Solve (6) by ADMM

**Input:** tensor data $\mathcal{X}$, parameter $\lambda$.
**Initialize:** $\mathcal{L}_0 = \mathcal{S}_0 = \mathcal{Y}_0 = 0$, $\rho = 1.1$, $\mu_0 = 1\mathrm{e}{-3}$, $\mu_{\max} = 1\mathrm{e}{10}$, $\epsilon = 1\mathrm{e}{-8}$.
**while** not converged **do**

1. Update $\mathcal{L}_{k+1}$ by
$$\mathcal{L}_{k+1} = \underset{\mathcal{L}}{\arg\min} \|\mathcal{L}\|_* + \frac{\mu_k}{2} \left\| \mathcal{L} + \mathcal{E}_k - \mathcal{X} + \frac{\mathcal{Y}_k}{\mu_k} \right\|_F^2;$$

2. Update $\mathcal{E}_{k+1}$ by
$$\mathcal{E}_{k+1} = \underset{\mathcal{E}}{\arg\min} \lambda\|\mathcal{E}\|_1 + \frac{\mu_k}{2} \left\| \mathcal{L}_{k+1} + \mathcal{E} - \mathcal{X} + \frac{\mathcal{Y}_k}{\mu_k} \right\|_F^2;$$

3. $\mathcal{Y}_{k+1} = \mathcal{Y}_k + \mu_k(\mathcal{L}_{k+1} + \mathcal{E}_{k+1} - \mathcal{X})$;

4. Update $\mu_{k+1}$ by $\mu_{k+1} = \min(\rho\mu_k, \mu_{\max})$;

5. Check the convergence conditions
$$\|\mathcal{L}_{k+1} - \mathcal{L}_k\|_\infty \leq \epsilon, \|\mathcal{E}_{k+1} - \mathcal{E}_k\|_\infty \leq \epsilon,$$
$$\|\mathcal{L}_{k+1} + \mathcal{E}_{k+1} - \mathcal{X}\|_\infty \leq \epsilon.$$

**end while**

### 4.1. Exact Recovery from Varying Fractions of Error

We first verify the correct recovery guarantee in Theorem 3.1 on random data with different fractions of error. For simplicity, we consider the tensors of size $n \times n \times n$, with varying dimension $n = 100, 200$ and $300$. We generate a $\text{rank}_\text{t}$-$r$ tensor $\mathcal{L}_0 = \mathcal{P} * \mathcal{Q}$, where the entries of $\mathcal{P} \in \mathbb{R}^{n \times r \times n}$ and $\mathcal{Q} \in \mathbb{R}^{r \times n \times n}$ are independently sampled from an $\mathcal{N}(0, 1/n)$ distribution. The support set $\Omega$ (with size $m$) of $\mathcal{S}_0$ is chosen uniformly at random. For all $(i, j, k) \in \Omega$, let $[\mathcal{S}_0]_{ijk} = [\mathcal{M}]_{ijk}$, where $\mathcal{M}$ is a tensor with independent Bernoulli $\pm 1$ entries.

We test on two settings. Table 1 (top) gives the results of the first scenario with setting $r = \text{rank}_\text{t}(\mathcal{L}_0) = 0.1n$ and $m = \|\mathcal{S}_0\|_0 = 0.1n^3$. Table 1 (bottom) gives the results for a more challenging scenario with $r = \text{rank}_\text{t}(\mathcal{L}_0) = 0.1n$ and $m = \|\mathcal{S}_0\|_0 = 0.2n^3$. It can be seen that our convex program (6) gives the correct rank estimation of $\mathcal{L}_0$ in all cases and also the relative errors $\|\hat{\mathcal{L}} - \mathcal{L}_0\|_F/\|\mathcal{L}_0\|_F$ are small, less than $10^{-5}$. The sparsity estimation of $\mathcal{S}_0$ is not exact as the rank estimation. The reason may be that the sparsity of errors is much larger than the rank of $\mathcal{L}_0$. But note that the relative errors $\|\hat{\mathcal{S}} - \mathcal{S}_0\|_F/\|\mathcal{S}_0\|_F$ are all small, less than $10^{-8}$ (actually much smaller than the relative errors of the recovered low-rank component). These results well verify the correct recovery phenomenon as claimed in Theorem 3.1 for (6) with the chosen $\lambda$ in theory.

### 4.2. Phase Transition in Rank and Sparsity

The results in Theorem 3.1 shows the perfect recovery for incoherent tensor with $\text{rank}_\text{t}(\mathcal{L}_0)$ on the order of $n/(\mu(\log nn_3)^2)$ and the sparsity of $\mathcal{S}_0$ on the order of $n^2 n_3$. Now we exam the recovery phenomenon with varying rank of $\mathcal{L}_0$ and varying sparsity of $\mathcal{S}_0$. We consider two sizes of $\mathcal{L}_0 \in \mathbb{R}^{n \times n \times n_3}$: (1) $n = 100, n_3 = 50$; (2) $n = 100, n_3 = 50$. We generate $\mathcal{L}_0 = \mathcal{P} * \mathcal{Q}$, where the entries of $\mathcal{P} \in \mathbb{R}^{n \times r \times n}$ and $\mathcal{Q} \in \mathbb{R}^{r \times n \times n}$ are independently sampled from an $\mathcal{N}(0, 1/n)$ distribution. For $\mathcal{S}_0$, we consider a Bernoulli model for its support and random signs for its values:

$$[\mathcal{S}_0]_{ijk} = \begin{cases} 1, & \text{w.p. } \rho_s/2, \\ 0, & \text{w.p. } 1 - \rho_s, \\ -1, & \text{w.p. } \rho_s/2. \end{cases} \quad (17)$$

We set $r/n$ as all the choices in $[0.01 : 0.01 : 0.5]$, and $\rho_s$ in $[0.01 : 0.01 : 0.5]$. For each $(r, \rho_s)$-pair, we simulate 10 test instances and declare a trial to be successful if the recovered $\hat{\mathcal{L}}$ satisfies $\|\hat{\mathcal{L}} - \mathcal{L}_0\|_F/\|\mathcal{L}_0\|_F \leq 10^{-3}$. Figure 2 plots the fraction of correct recovery for each pair (black = 0% and white = 100%). It can be seen that there is a large

Table 1: Correct recovery for random problems of varying size.

| $n$ | $r$ | $m$ | $\text{rank}_\text{t}(\hat{\mathcal{L}})$ | $\|\hat{\mathcal{S}}\|_0$ | $\frac{\|\hat{\mathcal{L}} - \mathcal{L}_0\|_F}{\|\mathcal{L}_0\|_F}$ | $\frac{\|\hat{\mathcal{S}} - \mathcal{S}_0\|_F}{\|\mathcal{S}_0\|_F}$ |
|---|---|---|---|---|---|---|
| \multicolumn{7}{c}{$r = \text{rank}_\text{t}(\mathcal{L}_0) = 0.1n, m = \|\mathcal{S}_0\|_0 = 0.1n^3$} |
| 100 | 10 | 1e5 | 10 | 101,952 | 4.8e−7 | 1.8e−9 |
| 200 | 20 | 8e5 | 20 | 815,804 | 4.9e−7 | 9.3e−10 |
| 300 | 30 | 27e5 | 30 | 2,761,566 | 1.3e−6 | 1.5e−9 |
| \multicolumn{7}{c}{$r = \text{rank}_\text{t}(\mathcal{L}_0) = 0.1n, m = \|\mathcal{S}_0\|_0 = 0.2n^3$} |
| 100 | 10 | 2e5 | 10 | 200,056 | 7.7e−7 | 4.1e−9 |
| 200 | 20 | 16e5 | 20 | 1,601,008 | 1.2e−6 | 3.1e−9 |
| 300 | 30 | 54e5 | 30 | 5,406,449 | 2.0e−6 | 3.5e−9 |

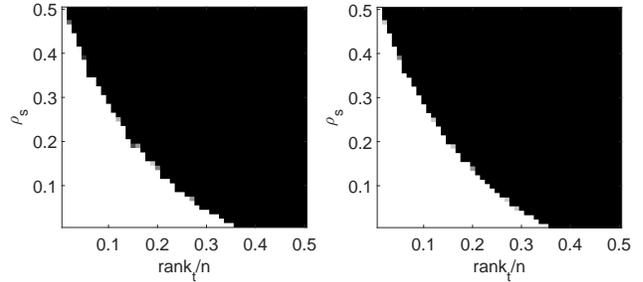

(a) $n_1 = n_2 = 100, n_3 = 50$    (b) $n_1 = n_2 = 200, n_3 = 50$

Figure 3: Correct recovery for varying rank and sparsity. Fraction of correct recoveries across 10 trials, as a function of $\text{rank}_\text{t}(\mathcal{L}_0)$ (x-axis) and sparsity of $\mathcal{S}_0$ (y-axis). The experiments are test on two different sizes of $\mathcal{L}_0 \in \mathbb{R}^{n_1 \times n_2 \times n_3}$.

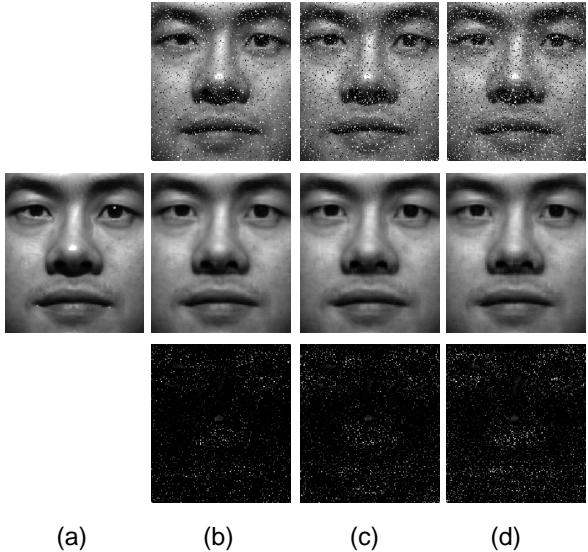

(a) (b) (c) (d)

Figure 4: Removing random noises from face images. (a) Original image; (b)-(d): Top: noisy images with 10%, 15% and 20% pixels corrupted; Middle: the low-rank component recovered by TRPCA; Bottom: the sparse component recovered by TRPCA.

region in which the recovery is correct. These results are quite similar as that in RPCA, see Figure 1 (a) in [4].

### 4.3. TRPCA for Image Recovery

We apply TRPCA for image recovery from the corrupted images with random noises. We will show that the recovery performance is still satisfied with the choice of $\lambda = 1/\sqrt{n_{(1)}n_3}$ on real data.

We conduct two experiments. The first one is to recover face images (of the same person) with random noises as that in [8]. Assume that we are given $n_3$ gray face images of size $h \times w$. Then we can construct a 3-way tensor $\mathcal{X} \in \mathbb{R}^{h \times w \times n_3}$, where each frontal slice is a face image[2]. An extreme situation is that all these $n_3$ face images are all the same. Then the tubal rank of $\mathcal{X}$ will be 1, which is very low. However, the real face images often violate such low-rank tensor assumption (the same observation for low-rank matrix assumption when the images are vectorized), due to different noises. Figure 4 (a) shows an example image taken from the Extended Yale B database [7]. Each subject of this dataset has 64 images, and each has resolution $192 \times 168$. We simply select 32 images with different illuminations per subject, and construct a 3-way tensor $\mathcal{X} \in \mathbb{R}^{192 \times 168 \times 32}$. Then, for each image, we randomly select a fraction of pixels to be corrupted with random pixel values. Then we solve TRPCA with $\lambda = 1/\sqrt{n_{(1)}n_3}$ to recover the face images.

[2]We also test TRPCA based on different ways of tensor data construction and find that the results are similar.

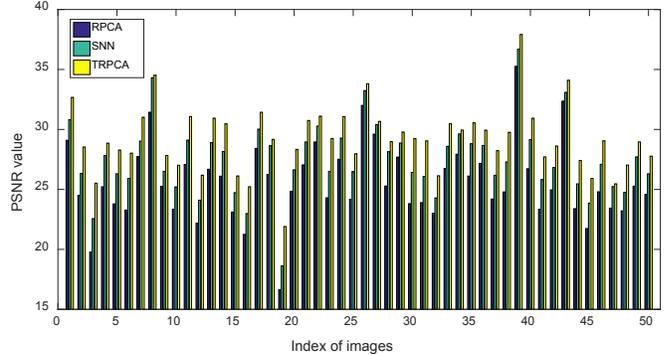

Figure 5: Comparison of the PSNR values of RPCA, SNN and TRPCA for image denoising on 50 images.

Figure (4) (b)-(d) shows the recovered images from different proportions of corruption. It can be seen that it successfully removes the random noises. This also verifies the effectiveness of our choice of $\lambda$ in theory.

The second experiment is to apply TRPCA for image denoising. Different from the above face recovery problem which has many samples of a same person, this experiment is tested on the color image with one sample of 3 channels. A color image of size $n_1 \times n_2$, is a 3-way tensor $\mathcal{X} \in \mathbb{R}^{n_1 \times n_2 \times 3}$ in nature. Each frontal slice of $\mathcal{X}$ is corresponding to a channel of the color image. Actually, each channel of a color image may not be of low-rank. But it is observed that their top singular values dominate the main information. Thus, the image can be approximately recovered by a low-rank matrix [17]. When regarding a color image as a tensor, it can be also well reconstructed by a low tubal rank tensor. The reason is that t-SVD is capable for compression as SVD, see Theorem 4.3 in [14]. So we can apply TRPCA for image denoising. We compare our method with RPCA and the SNN model (4) [10] which also own the theoretical guarantee.

We randomly select 50 color images from the Berkeley Segmentation Dataset [20] for the test. For each image, we randomly set 10% of pixels to random values in $[0, 255]$. Note that this is a challenging problem since at most there are 30% of pixels corrupted and the positions of the corrupted pixels are unknown. We use our TRPCA with $\lambda = 1/\sqrt{n_{(1)}n_3}$. For RPCA (1), we apply it on each channel independently with $\lambda = 1/\sqrt{n_{(1)}}$. For SNN (4), we find that its performance is very bad when $\lambda_i$'s are set to the values suggested in theory [10]. We empirically set $\lambda_1 = \lambda_2 = 15$ and $\lambda_3 = 1.5$ which make SNN perform well in most cases. For the recovered image, we evaluate its quality by the Peak Signal-to-Noise Ratio (PSNR) value. The higher PSNR value indicates better recovery performance. Figure 5 shows the PSNR values of the compared methods on all 50 images. Some examples are shown in Figure 6. It can be seen that our TRPCA obtains the best recovery performance. The two tensor methods, TRPCA and SNN, also perform much better than RPCA. The reason is that RPCA, which performs the matrix recovery on each channel independently, is not able to use the information across channels, while the tensor methods improve the performance by taking the advantage of the multi-dimensional structure of data.

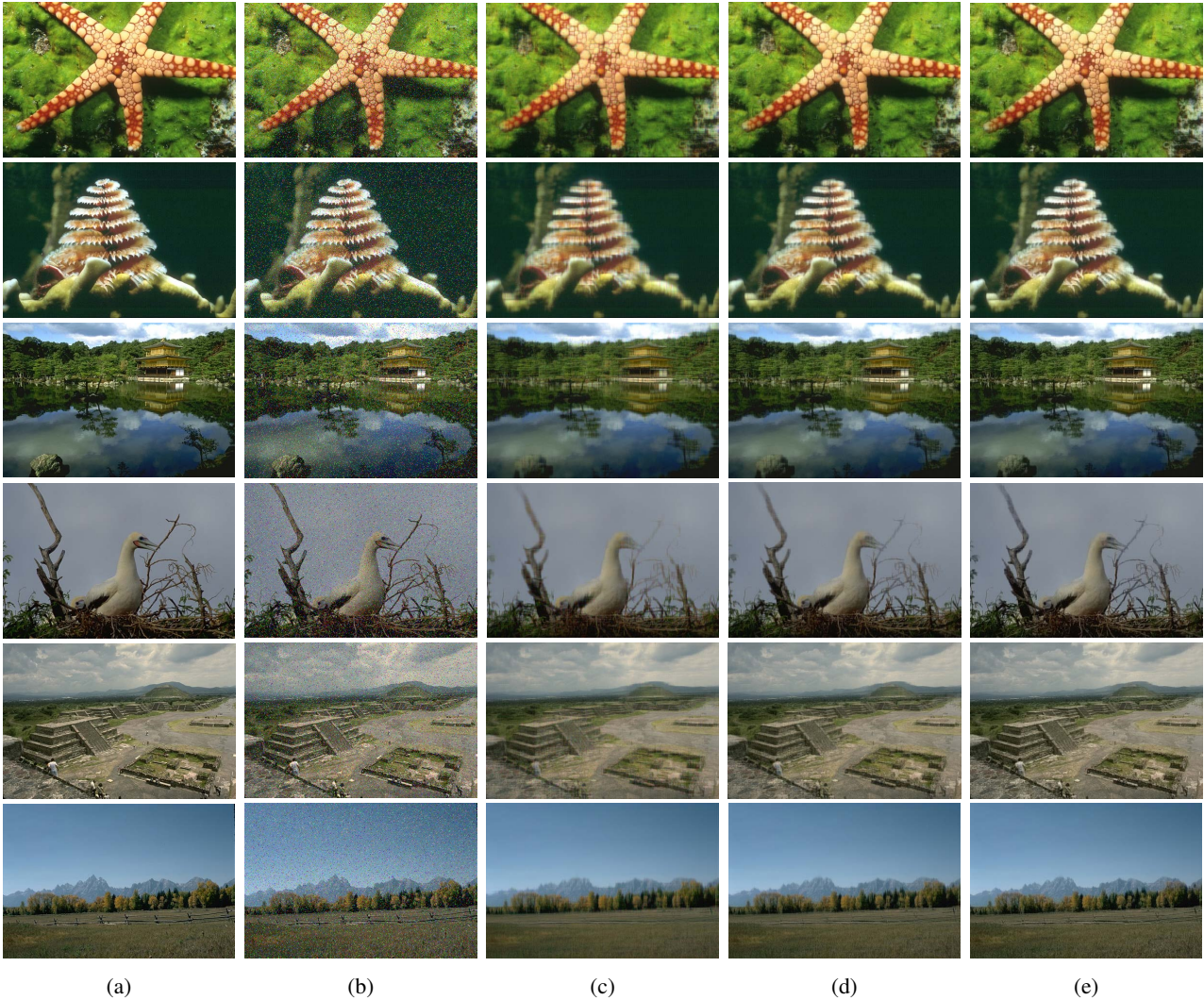

Figure 6: Comparison of Image recovery. (a) Original image; (b) Noisy image; (c)-(e) Recovered images by RPCA, SNN and TRPCA, respectively. **Best viewed in ×2 sized color pdf file.**

## 5. Conclusions and Future Work

In this work, we study the Tensor Robust Principal Component (TRPCA) problem which aims to recover a low tubal rank tensor and a sparse tensor from their sum. We show that the exact recovery can be obtained by solving a tractable convex program which does not have any free parameter. We establish a theoretical bound for the exact recovery as RPCA. Benefit from the "good" property of t-SVD, both our model and theoretical guarantee are natural extension of RPCA. Numerical experiments verify our theory and the applications for image denoising shows its superiority over previous methods.

This work verifies the remarkable ability of convex optimizations for low-rank tensors and sparse errors recovery. This suggests to use these tools of tensor analysis for other applications, e.g., image/video processing, web data analysis, and bioinformatics. Also, consider that the real data usually are of high dimension, the computational cost will be high. Thus developing the fast solver for low-rank tensor analysis is an important direction. It is also interesting to consider nonconvex low-rank tensor models [17, 18].

## Acknowledgements

This research is supported by the Singapore National Research Foundation under its International Research Centre@Singapore Funding Initiative and administered by the IDM Programme Office. J. Feng is supported by NUS startup grant R263000C08133. Z. Lin is supported by China 973 Program (grant no. 2015CB352502), NSF China (grant nos. 61272341 and 61231002), and MSRA.